\documentclass[twoside,11pt]{article}
\usepackage{jmlr2e}

\usepackage{amsmath}
\usepackage{amssymb}  
\usepackage{float}
\usepackage[table]{xcolor}
\usepackage{graphicx}
\usepackage{multirow}
\usepackage{enumitem}
\usepackage[table]{xcolor}
\usepackage{url}
\usepackage{subfigure} 
\usepackage{algorithm}
\usepackage{algpseudocode}
\usepackage{listings}
\usepackage{hhline}
\usepackage{pbox}

\algnewcommand\algorithmicinput{\textbf{Input:}}
\algnewcommand\INPUT{\item[\algorithmicinput]}

\algnewcommand\algorithmicoutput{\textbf{Output:}}
\algnewcommand\OUTPUT{\item[\algorithmicoutput]}

\ShortHeadings{Uncovering Voice Misuse Using Symbolic Mismatch}{Ghassemi, Marzyeh et al.}
\firstpageno{1}

\begin{document}

\title{Uncovering Voice Misuse Using Symbolic Mismatch}

\author{\name Marzyeh Ghassemi \email mghassem@mit.edu \\
    \addr Computer Science and Artificial Intelligence Laboratory\\
   	Massachusetts Institute of Technology, Cambridge, MA\\  
    \AND
	\name Zeeshan Syed \email zhs@umich.edu \\
    \addr University of Michigan \\
    \addr Ann Arbor, MI \\       
	\AND
    \name Daryush D. Mehta \email mehta.daryush@mgh.harvard.edu \\
    \addr MGH Institute of Health Professions, Boston, MA \\
	Massachusetts General Hospital, Boston \\
	Harvard Medical School, Boston, MA \\
	\AND
	\name Jarrad H. Van Stan \email JVanStan@mghihp.edu \\
    \addr MGH Institute of Health Professions, Boston, MA \\
	Massachusetts General Hospital, Boston \\
    \AND
	\name Robert E. Hillman \email rhillman@partners.org \\
    \addr MGH Institute of Health Professions, Boston, MA \\
	Massachusetts General Hospital, Boston \\
	Harvard Medical School, Boston, MA \\
    \AND
	\name John V. Guttag \email guttag@mit.edu \\
    \addr Computer Science and Artificial Intelligence Laboratory\\
   	Massachusetts Institute of Technology, Cambridge, MA\\}

\maketitle
\begin{abstract}
Voice disorders affect an estimated 14 million working-aged Americans, and many more worldwide. We present the first large scale study of vocal misuse based on long-term ambulatory data collected by an accelerometer placed on the neck. We investigate an unsupervised data mining approach to uncovering latent information about voice misuse.

We segment signals from over 253 days of data from 22 subjects into over a hundred million single glottal pulses (closures of the vocal folds), cluster segments into symbols, and use symbolic mismatch to uncover differences between patients and matched controls, and between patients pre- and post-treatment. Our results show significant behavioral differences between patients and controls, as well as between some pre- and post-treatment patients. Our proposed approach provides an objective basis for helping diagnose behavioral voice disorders, and is a first step towards a more data-driven understanding of the impact of voice therapy.
\end{abstract}

\section{Introduction}
An estimated 7\% of the working-age population in the U.S. is affected by a voice disorder~\citep{roy2005voice,oecd}. Most cases of voice disorders result from vocal misuse (exerting excessive muscle force or physical effort while vocalizing). This is typically referred to as {\it vocal hyperfunction}. In some patients vocal hyperfunction causes a deterioration in voice quality and vocal fatigue but without any underlying tissue pathology; this is commonly referred to as \textit{muscle tension dysphonia} (MTD). Unlike those with vocal fold pathology (e.g. nodules or polyps), MTD patients are notoriously difficult to characterize because there is no consensus on an objective biomarker. Previous studies have also demonstrated that commonly held ``indicators'' of MTD appear frequently in individuals that have no known voice disorder~\citep{stager2003incidence,behrman2003anterior}.

Because MTD is behaviorally induced, treatment typically involves an attempt to modify vocal behavior through speech/voice therapy~\citep{hsiung2003characteristic}. 
However, MTD can be manifested in a wide range of maladaptive vocal behaviors (e.g., various degrees of strain or breathiness) whose nature and severity can display significant situational variation (e.g., variation associated with changes in levels of stress~\citep{demmink2008neurovegetative}). Clinicians currently rely on patient self-reporting and self-monitoring to assess the prevalence and persistence of these behaviors during diagnosis and management. But these reports are highly subjective and are known to be unreliable. 

The work reported here is part of an ongoing project to gain insight into the complex relationships underlying vocal hyperfunction by analyzing data collected from an accelerometer (ACC) placed on the neck~\citep{mehta2012mobile}. We use an accelerometer rather than an acoustic microphone to protect the privacy of subjects. Recent studies have demonstrated some success applying supervised learning to ACC data to distinguish between patients with and without existing vocal fold pathology~\citep{ghassemi2014learning}. The work reported on here is more challenging in three respects:
\begin{itemize}[noitemsep,nolistsep]
\item Patients with MTD have a behavioral disorder whereby they misuse their vocal folds, but do not have an anatomical abnormality. Therefore their voices are sometimes abnormal and sometimes not.
\item While it is possible to obtain subjective expert-generated labels for acoustic recordings, it is impossible to obtain labels at the level of individual utterances for hundreds of millions of utterances. Additionally, even if someone were willing to devote the time to labeling a substantial number of utterances, the mapping between the ACC signal and voice misuse is not currently known. Consequently, there is no opportunity to use supervised learning to classify utterances.
\item Rather than attempting to classify individual subjects, we attempt to uncover the key differences between many kinds of intermittently occurring hyperfunctional and normal voice use--without prior knowledge of what characterizes such behaviors.
\end{itemize} 

We attack the problem of quantifying vocal hyperfunction by clustering glottal pulses using symbolic mismatch~\citep{syed2011unsupervised} - a technique previously used to study ECG signals. We segmented over 110 million glottal pulses from the ACC signals for subjects, and then clustered them into symbols. We then used symbolic mismatch to compare the frequencies and shapes of those symbols between subjects, leading to a distance measure between each pair of subjects. Finally, based on this distance measure, we clustered subject-days.

To evaluate our approach, we used 253 subject-days of data obtained from 11 patients and 11 matched controls (\textit{Control}). Data from patients was gathered both before they underwent voice therapy ({\it PreTx}) and after voice therapy ({\it PostTx}). Though we know that each individual exhibits different vocal behaviors within a day, we hypothesized that subject-class-specific differences in the distribution of the behaviors would be reflected in the distribution of subject days in each cluster. To check this we calculated a total concentration measure based on the density of each class of subject in each cluster.

We demonstrate that our approach separates subject-days from PreTx/Control subjects into clusters with a total concentration measure of 0.70. This result was statistically significant when compared to clusterings from randomly drawn distances (p $<$ 0.001). We also demonstrate that subject-days from PostTx and Control subjects are closer to one another than subject-days from PreTx and Control subjects. Moreover, the difference between PostTx and Control subject-days was not statistically significant. These last two results suggest that, on average, therapy moves the behavior of patients towards ``normal.''

We summarize our contributions as follows:
\begin{itemize}[noitemsep,nolistsep]
\item We are the first paper to use continuous data from accelerometers placed on the necks of patients and matched controls in an ambulatory setting to uncover latent information about voice misuse. 

\item We present and apply a fully unsupervised learning method to over a hundred million single glottal pulses from 253 days of data, and quantitatively evaluated the results. 

\item We were able to uncover significant behavioral differences between patients and controls, as well as between some pre- and post-treatment patients. 

\item We believe that our approach can be used as an objective basis for helping diagnose behaviorally-based voice disorders, as a first step towards a more empirical understanding of the impact of voice therapy, and eventually to help design biofeedback tools that could assist patients in avoiding damaging vocal behaviors.
\end{itemize}

\section{Background}
Many common voice disorders are believed to be caused by abusive vocal behaviors, generically called vocal hyperfunction. This voice misuse is assessed using patients' self-reporting, which is notoriously inaccurate~\citep{buekers1995vocal,rantala1999relationship,ohlsson1989voice}. Voice disorders caused by hyperfunction can have a devastating impact on an individual's ability to speak and sing. It has been previously observed that some patients with vocal hyperfunction develop vocal pathology such as nodules, but others develop vocal fatigue without tissue changes. This has resulted in two categories of vocal hyperfunction: adducted (associated with the development of nodules and polyps) or non-adducted (no development of tissue pathology). In this work, our goal was to determine if specific patterns of glottal pulses were associated with non-adducted hyperfunction.

\begin{figure}[ht]
\centering
\includegraphics[width=0.5\columnwidth]{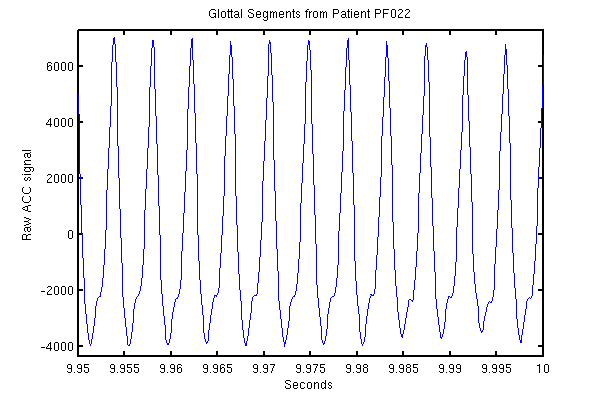}
\caption{A sustained vowel ``a'', containing 10 peak-to-peak glottal pulses in 0.05 seconds.}
\label{fig:accExamples}
\end{figure}

Devices that use a neck-placed miniature accelerometer (ACC) as a phonation sensor have shown potential for accurate, unobtrusive, and privacy-preserving long-term monitoring of vocal function~\citep{mehta2012mobile} (Figure \ref{fig:accExamples}). The individual periods (pulses) in  the ACC signal have a general shape that reflects the vibratory pattern of the vocal folds during phonation, and vary with changes in vocal function/quality. Recently, researchers have examined vocal hyperfunction using summary features obtained from ambulatory monitoring~\citep{roy2013evidence,ghassemi2014learning}, but these assessments were based on aggregates, and were not designed to detect periods of hyperfunction. Glottal pulses obtained from the ACC signal have a general shape that describes the acceleration of the vocal folds as they vibrate to create airflow for voice production. Because ACC signals have only recently become available, variations in the segmented pulses are not currently well-characterized.  
\vspace{-2.0em}

\section{Methods}
To generate symbols for every subject-day tuple, we segmented each daily ACC signal into non-overlapping frames to create a set of variable-length, peak-to-peak glottal pulse segments. We then computed the pulse-to-pulse distance using a lower bounds to dynamic time warping (DTW) distance, and created clusters iteratively as described below.

\subsection{Glottal Pulse Symbolization}
\paragraph{Segmentation} We begin with the continuous univariate timeseries of a single subject's ACC recording on a given day (a ``subject-day''). This signal ${\bf x} \in R^T$ is a collection of $T$ samples, i.e. $\textbf{x} = \{x_1, ..., x_t, ..., x_T\}$, in which measurements are regularly-sampled. We split the ACC signals into individual glottal pulses by detecting characteristic peaks. Peak detection involved 1) using an off-the-shelf peak detection algorithm~\citep{matlabSigProcToolbox} to make a first guess at peak locations based on amplitude, and 2) using an estimate of the subject's underlying vocal pitch to correct missing and spurious peaks. After segmentation, we have a vector of $M$ daily glottal pulse segments, $\textbf{x}_{seg} = \{(x_{t_1}, ..., x_{t_2}), ..., (x_{t_{2M-1}}, ..., x_{t_{2M}})\}$, where $t_1, ..., t_{2M}$ are increasing but not necessarily contiguous, so that $0 \leq t_1 \leq ... \leq t_{2M} \leq T$. Notationally, we re-label this as $\textbf{x}_{seg} = \{z_1, ..., z_M\}$ where $z_1 = (x_{t_1}, ..., x_{t_2})$, $z_M = (x_{t_{2M-1}}, ..., x_{t_{2M}})$. 

The amplitude of each glottal pulse was scaled to units of sound pressure decibels (dbSPL) based on an estimated linear fitting between ACC signal units and average dbSPL for the subject on that day to determine periods of voicing. The length of each individual segmented pulse varied; to compare all pulses, we length-normalized pulses by evenly up-sampling all segments to the longest segment length. 

\paragraph{Pulse-Pulse Distance Computation} Silent segments were grouped by their length into bins of 1 second, 1 minute, 10 minutes, and an hour or more.\footnotemark
\footnotetext{A lot of any subject's day is spent in silence; the amount varied from 86\%-95\%. The mean number of voiced pulses per patient was 3,427,367.} 
To account for the large variation in subjects' pattern of voice use across days (e.g., teachers typically spoke less on weekends), we chose to examine each day separately. For each subject-day, we start with the constructed vector $\textbf{x}_{seg} = \{z_1, ..., z_M\}$ and compute the distance between all pulses $z_i$ and $z_j$ using the Keogh Bounds (LB\_Keogh) \citep{wang2013experimental} as a surrogate for DTW. LB\_Keogh is a tight lower bounds to DTW between a candidate signal $C$ and query signal $Q$, and is considerably more computationally efficient than DTW. 

\paragraph{Symbolization for Symbolic Feature Creation} We next used hierarchical clustering with Ward's linkage, which minimizes the total within-cluster variance, to cluster a randomly selected initial subsample of 3,000 pulses per subject-day. We used a distance cutoff of 30\% of the maximum distance to determine $k$, the number of clusters. Having chosen $k$, we then used iterative k-means to cluster all of the pulses $z_1, ..., z_M$. Each of the $k$ clusters can be considered as representing a class of glottal pulses whose members have a similar shape. We label each of these classes with its centroid, and create a vector of length $k$ of symbolic features $\textbf{v}$ for each subject-day, where $\textbf{v} = \{(s_1, f_1) ..., (s_k, f_k)\}$, $s_i$ is the $i^{th}$ class centroid, and $f_i = \frac{|s_i|}{\sum_j |s_j|}$.
In creating $\textbf{v}$, we have now abstracted from a stream of millions of glottal pulses into an finite alphabet of symbols with matching frequencies of occurrence.

\paragraph{Symbolic Mismatch Distance Measure}
Once symbolic features $\textbf{v}$ were created for each subject-day, we defined the overall distance measure between each pair of $\textbf{v}$'s as the symbolic mismatch distance $D_{mismatch}$[i,j]. For subject-days $\textbf{v}_i$ and $\textbf{v}_j$, $D_{mismatch}$[i,j] is the aggregate sum of the weighted distance between class centroids. 
\begin{algorithm}
\caption{Symbolic mismatch calculation between subject/day tuple pairs.}
\label{lst:mismatch}
\begin{algorithmic}[1]
\INPUT{Transformed data from subject/day tuples $v_i$ and $v_j$}
\OUTPUT{Weighted distance between $v_i$ and $v_j$}
\State{initialize W $\gets$ 0}
\For{each $s_a$ $\in$ $v_i$}
	\For{each $s_b$ $\in$ $v_j$}
		\State{ W $\gets$ W + $f_a$ * $f_b$ * LB\_Keogh($s_a$, $s_b$) }
	\EndFor
\EndFor
\State{$D_{mismatch}$[i, j] $\gets$ W}
\end{algorithmic}
\end{algorithm}

\subsection{Subject-Day Clustering and Evaluation}
\label{sec:eval}
\label{sec:concmeas}
We evaluate a clustering of $Q$ subject-days $\textbf{v}_1, ..., \textbf{v}_Q$ across $n$ clusters in two ways: \textit{class concentration} and \textit{subject concentration}. For an individual cluster $\textbf{c}$ with some number of total (subject-day, class label) pairs, i.e. suppose there are $o$ pairs of them $\textbf{c} = \{(\textbf{v}_{1}, \textbf{l}_{1}), ..., (\textbf{v}_o, \textbf{l}_{o})\}$, class concentration is the cluster's ratio of the dominant label to the total number of in-cluster subject-days. Subject concentration is calculated similarly, but we count $\textbf{v}$ from the same subject only once. For example, suppose we have a cluster with items $c_1 = \{(v_{1-1}, 0), (v_{2-1}, 1), (v_{2-3}, 1), (v_{3-1}, 1), (v_{3-5}, 1)\}$\footnote{Corresponding to subject 1-day 1 with label 0, subject 2-day 2 and subject 2-day 3 labeled 1, etc.}, the class concentration would be $conc_{class} = \frac{4}{1+4}$ and the subject concentration would be $conc_{subj} = \frac{2}{2+1}$. 

\paragraph{Total Concentration} We define the total concentration for both metrics across clusters as the weighted sum of all individual cluster concentrations. Specifically, for $n$ clusters $c_1 \ldots c_n$ with concentrations $h_1 \ldots h_n$, total concentration is defined as $total\_conc = \sum_{i = 1}^n h_i * |c_i|$. Note that when there are two classes, the total concentration can range from $[0.5,1]$, since the least concentrated cluster possible is 0.5. To check statistical significance, we tested the null hypothesis that the groupings obtained with $D_{mismatch}$ were different from a total concentration measure using random distances. We first define a random distance metric (RRDM) by sampling random values uniformly as $RRDM[i,j] = \mathcal{U} ([0, max\left\{D_{mismatch}\right\}])$, where $max \left \{ D_{mismatch} \right \}$ is the maximum distance seen from the actual symbolic mismatch. We sampled distances for each subject/tuple pair $\textbf{v}_i$ and $\textbf{v}_j$ 5,000 times, and cluster those (random) values. 
We clustered the RRDM values to obtain a distribution of total class concentration measures, fit an empirical CDF (ECDF) to these values, and computed the probability ($p$) of a total class concentration value greater than or equal to ours by chance ($1-ECDF(conc_{class}(D_{mismatch})))$. 

\section{Experiments}
\subsection{Data}
We considered 11 MTD patients with matched controls --- a total of 22 subjects. Diagnoses were based on evaluation by a laryngologist and speech-language pathologist. 
All patients were treated with behavioral voice therapy, and each patient was recorded for a minimum of six days both before and after undergoing treatment. This created a set of three categories in our data:
\begin{itemize}[noitemsep,nolistsep]
\item 11 pre-treatment MTD patients \textbf{(PreTx)},
\item the same 11 patients after behavioral voice therapy \textbf{(PostTx)}, and
\item 11 control subjects matched for age, gender, and occupation \textbf{(Control)}.
\end{itemize}
We used a neck-placed miniature accelerometer as a voice sensor and a smart phone as the data acquisition platform~\citep{mehta2012mobile}. The raw accelerometer signal was collected at 11,025 Hz, 16-bit quantization, and 80-dB dynamic range in order to obtain neck skin vibrations at frequencies up to 4,000 Hz. Our dataset contains 253 subject-days, corresponding to over 110 million segmented pulses (details in Appendix A). Working with a continuous ACC signal for each subject over the course of 7+ days yielded approximately 15 GB of data per subject. 

\subsection{Clinical Significance}
We investigated the utility of our method in addressing three clinical questions: 
\begin{enumerate}[noitemsep,nolistsep]
 	\item \textbf{Can our features be used to diagnose MTD (PreTx vs. Control subject/days)?}
    To address the first question, we performed an inter-subject comparison on PreTx vs. Control subjects, where we clustered all pre-therapy subject-days and all control subject-days. We did not expect a clean separation of all PreTx days from Control days to occur, because many MTD patients have ``good'' days where their voice use is like that of a vocally normal individual. Instead, our objective was to determine if a clustering of $D_{mismatch}$ could achieve a high concentration in the PreTx vs. Control comparison ($conc_{class}(PreTx/Con)$) that was significantly different from those that could be obtained by chance.

	\item \textbf{Can we detect a treatment effect (paired PreTx vs. PostTx subject/days)?}
To address the second question, we perform an intra-patient comparison on PreTx vs. PostTx subjects where we performed clusterings on a patient-patient basis, (i.e., we clustered all days, both pre and post treatment, on a patient-by-patient basis). 
    
 	\item \textbf{If our features can be used to detect treatment effect, is the effect to move patients towards ``normal'' (PostTx vs. Control subject/days)?}
To address this question, we performed an inter-subject clustering on the PostTx vs. Control subjects, clustering all post-therapy subject-days and all control subject-days. Our objective was to determine if this clustering would produce concentrations ($conc_{class}(PostTx/Con)$) which were not significantly different from those that could be obtained by chance. This would indicate that patients are difficult to distinguish from controls after they receive voice therapy.
\end{enumerate}

\subsection{Baseline Methods}
Our symbolic features (SF) were compared over subject-days using symbolic mismatch to generate a paired distance matrix, and the mismatch distance was clustered using hierarchical clustering and Ward's linkage. We compared clusterings generated from our method to clusterings from features generated by a recently proposed system for identifying adducted hyperfunctional patients versus their matched controls~\citep{ghassemi2014learning}. 

As in ~\citep{ghassemi2014learning}, we windowed the regularly sampled $\textbf{x} = \{x_1, ..., x_t, ..., x_T\}$ ACC signal into five-minute windows, computed the phonation frequency (f0) and acoustic sound pressure level (SPL) of non-overlapping 50 millisecond frames within each window (i.e., 6000 frames per window), and extracted statistical features of these acoustically inspired measures (e.g., the mean, skew, $5^{th}$ percentile value, etc.). Each subject-day is a feature matrix, where the number of features varied based on the amount of phonation in each subject-day. We also removed the most correlated features, yielding a total of 22 features. Once generated for each subject-day, these generate a Vector of Acoustic Features (VAF) that has multiple features summarizing a given subject-day tuple. We clustered VAF vectors from all subject-day tuples using k-means clustering with a squared Euclidean distance function.

While the VAF previously detected constantly-present pathology in adducted patients, we theorized they would create many incorrectly labeled windows for clustering in the periodically hyperfunctional MTD population. To address this, we took the feature-wise mean over all five-minute windows for a single subject-day, to obtain Mean Acoustic Features (MAF). These vectors were clustered with hierarchical clustering and Ward's linkage. 

We measured the total concentration in all clusterings as described in \ref{sec:concmeas}. For inter-subject comparisons, we investigated the sensitivity of our method and the baselines by varying the number of clusters in the final grouping ($n$) from 2 to 40; for the intra-subject comparisons we varied $n$ from 2 to 10. 

\section{Results}
\subsection{Control vs. PreTx Subjects - Potential for ambulatory screening tool}
\label{sec:precon}
\begin{figure}[ht]
\centering
\subfigure[Clustering of PreTx vs. Control subject-days]{%
\includegraphics[width=0.5\columnwidth]{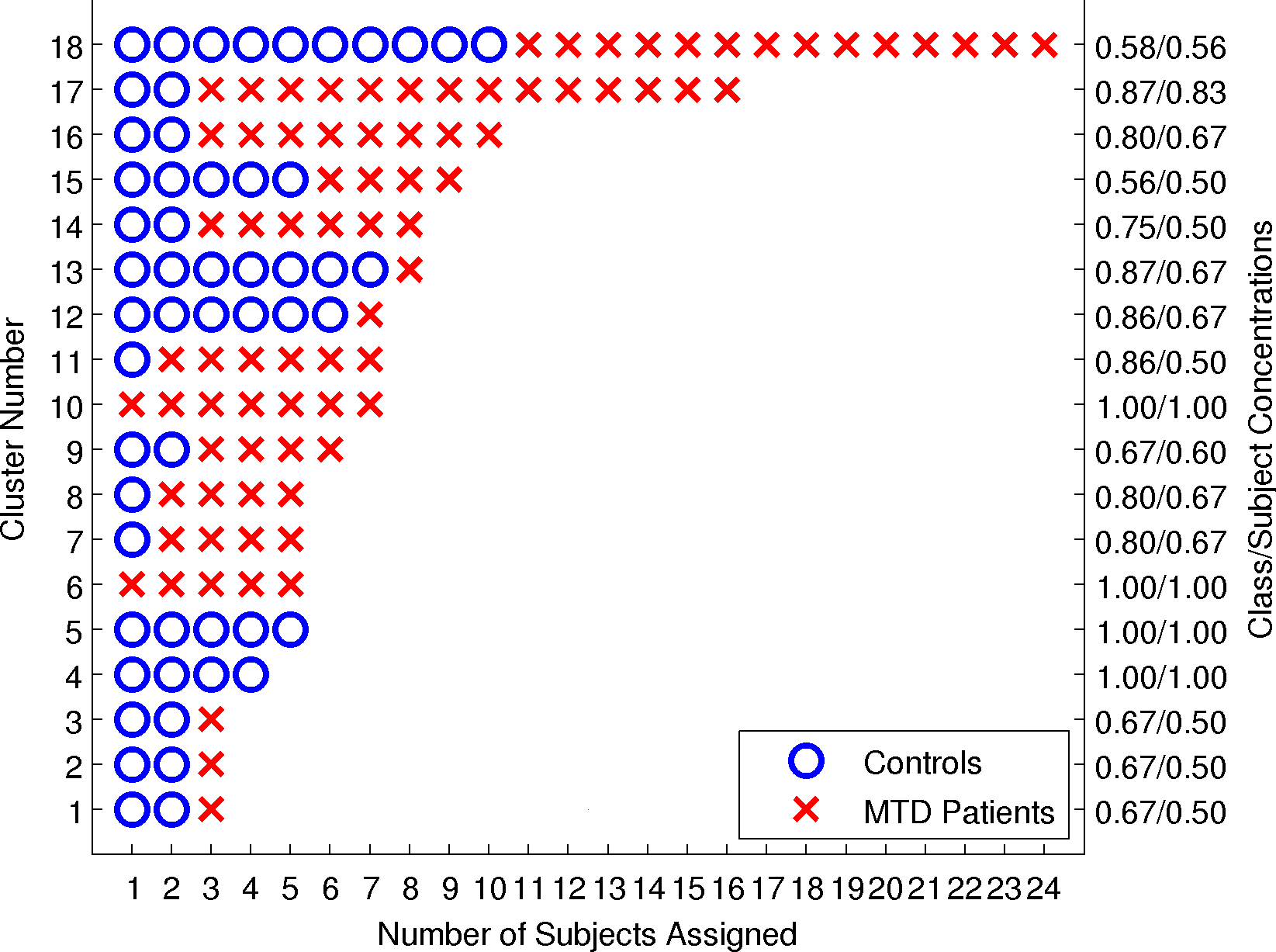}
\label{fig:pretxCon}}
\quad
\subfigure[Result vs. RRDM ECDF]{%
\includegraphics[width=0.25\columnwidth]{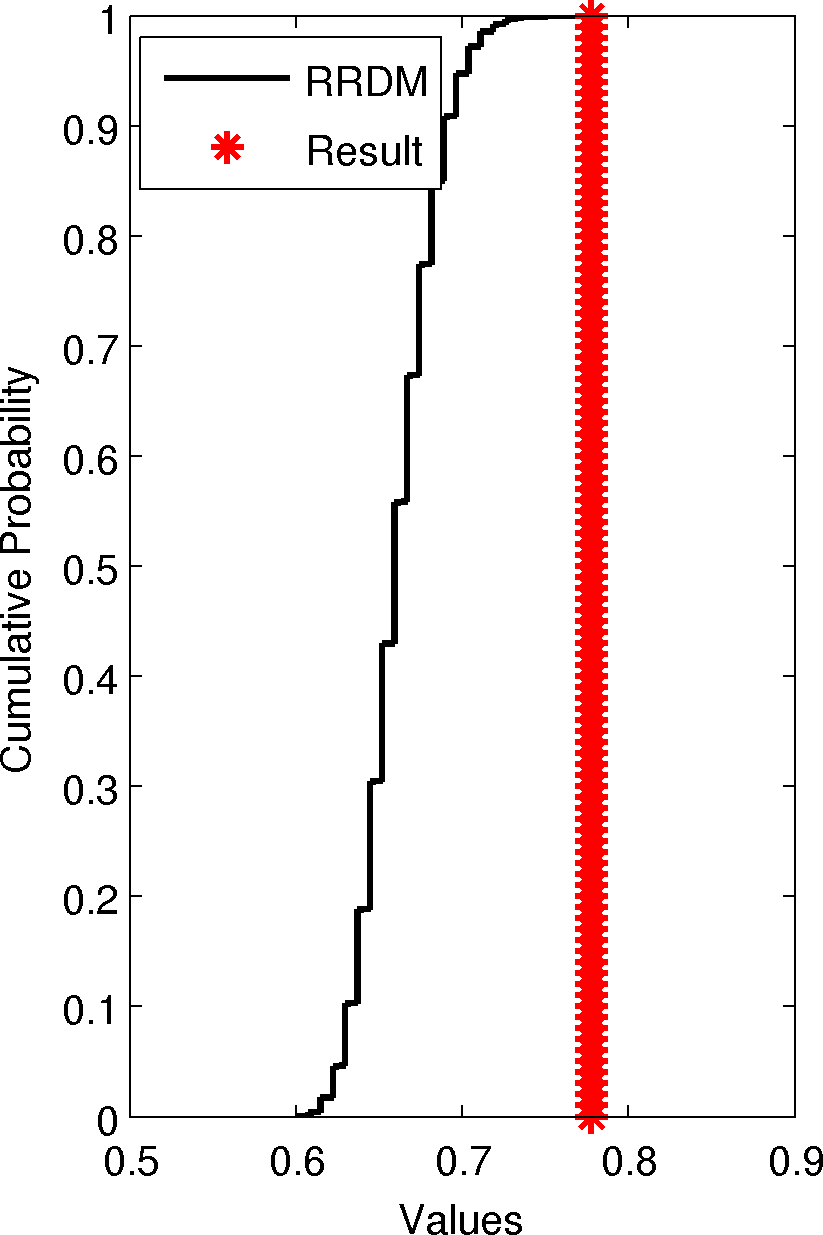}
\label{fig:ecdf1_2}}
\caption{We show a) the results of symbolic mismatch clustering of the control subject-days versus the PreTx subject-days (18 clusters, class concentration = 0.70) and b) the empirical CDF of the 5,000 RRDM clusterings versus our experimental results (p $=$ 0.001). As shown in b), Controls and PreTx patients were significantly different.} 
\label{fig:preconclustFigure}
\end{figure}

After performing clustering on all subject-day pairs from Control and PreTx subjects into 18 clusters, we obtained a total class concentration measure of 0.70. As shown in Figure \ref{fig:preconclustFigure}, using the RRDM clustering comparison, the difference between the PreTx and Control groups were statistically significant at p $<$ 0.001. There were a total of 135 subject/days in the comparison, and no cluster had data from only a single subject (total subject concentration measure of 0.65). Given the intermittent nature of voice misuse, it is reasonable that some days from PreTx patients cluster with Controls.
\vspace{-1.0em}

\subsection{PreTx vs. PostTx Subjects - Vocal therapy effect in pairs}
\label{sec:prepost}
We investigated if voice therapy had an effect that could be detected in our framework by using an intra-subject comparison on a patient-patient basis, so that all days from a patient pre-treatment were compared all days from the same patient post-treatment. As shown in Table \ref{tab:pairedRes}, the results vary for each patient, with some demonstrating more post-therapy differences than others. 
One possible explanation for a smaller intra-subject concentration is that improved vocal behavior for a particular subject was observable during a smaller time scale than we examined (e.g., better behavior during their evenings).
\vspace{-1.0em}
\begin{table}[htb]
\caption{Total concentration of per-patient PreTx vs. PostTx with three clusters. Concentrations that passed the empirical RRDM significance of p $<$ 0.01 are highlighted with **, and those with p $<$ 0.05 are marked with *.}
\label{tab:pairedRes}
\centering 
\begin{tabular}
{p{1cm}|p{1cm}|p{1cm}|p{1cm}|p{1cm}|p{1cm}|p{1cm}|p{1cm}|p{1cm}|p{1cm}|p{0.9cm}}
\hline
F023 & F027 & F040 & F048 & F052 & F064 & F069 & F071 & F100 & M035 & M074\\
\hline
0.73 & 0.65 & 0.81* & 1.0** & 0.63 & 0.69 & 0.67 & 1.0** & 0.86* & 0.57 & 0.79 \\
\hline
\end{tabular}
\end{table}

\subsection{Control vs. PostTx Subjects - Therapy moves subjects toward ``normal''}
\label{sec:postcon}
\begin{figure}[htb]
\centering
\subfigure[Clustering of PostTx vs. Control subject-days]{%
\includegraphics[width=0.5\columnwidth]{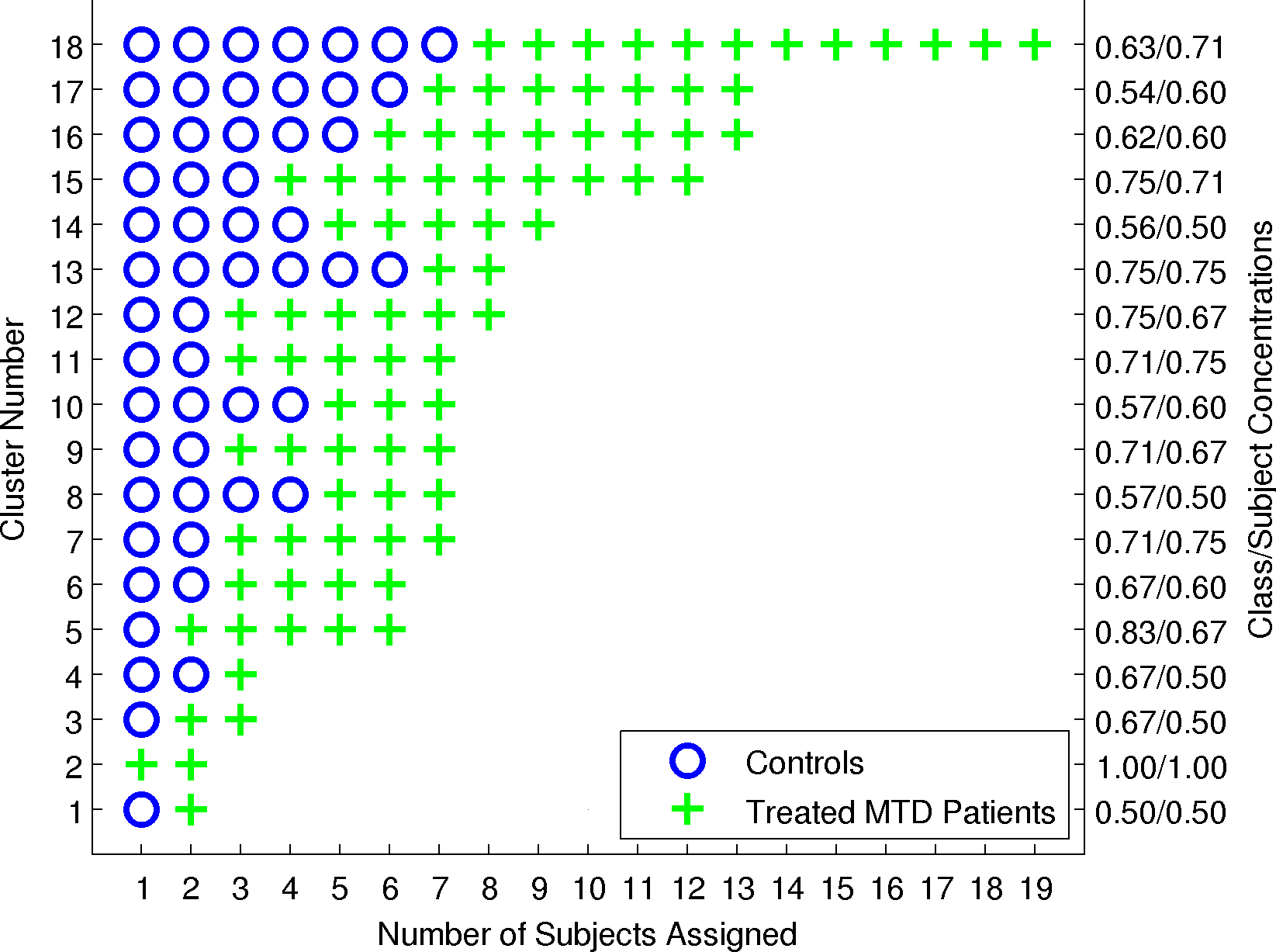}
\label{fig:posttxCon}}
\quad
\subfigure[Result vs. RRDM ECDF]{%
\includegraphics[width=0.25\columnwidth]{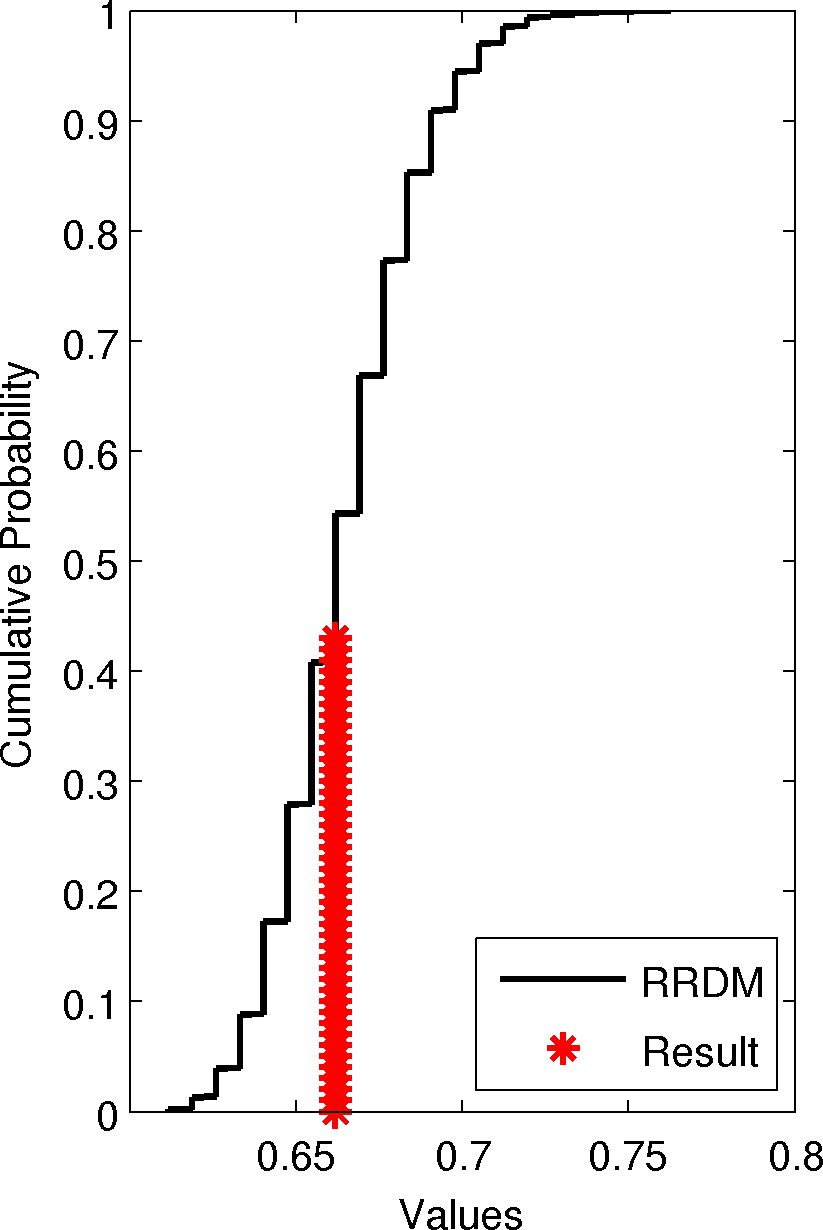}
\label{fig:ecdf2_2}}
\caption{We show a) the results of symbolic mismatch clustering of the control subject-days versus the PostTx subject-days and b) the empirical CDF of 5,000 random distance clusterings versus our experimental results. PostTx subject-days were not significantly different from the control group, suggesting that voice therapy does indeed move patients toward vocal normalcy.}
\label{fig:postconclustFigure}
\end{figure}
As shown in Figure \ref{fig:postconclustFigure}, after clustering the PostTx patients and Control subjects, we obtained a total class concentration of 0.63, and a subject concentration of 0.60. There was no statistically significant difference between these clustering and clusterings of the RRDM distances (p = 0.56). In this clustering of the 139 total days, PostTx patients only enrich a few clusters, and many clusters are evenly class-balanced. This suggests that our method is picking up changes caused by voice therapy, and that these changes are in the right direction.

\subsection{Sensitivity Analysis of Clustering Across Baselines and Clusters}
After successfully demonstrating differences in PreTx vs. Control subjects-days, and showing that PostTx subject-days are like those of the Controls, we examined the ability of our symbolic features (SF) to perform under varying numbers of clusters as compared to other methods (VAF and MAF). 

\begin{figure}[ht]
\centering
\includegraphics[width=0.6\columnwidth]{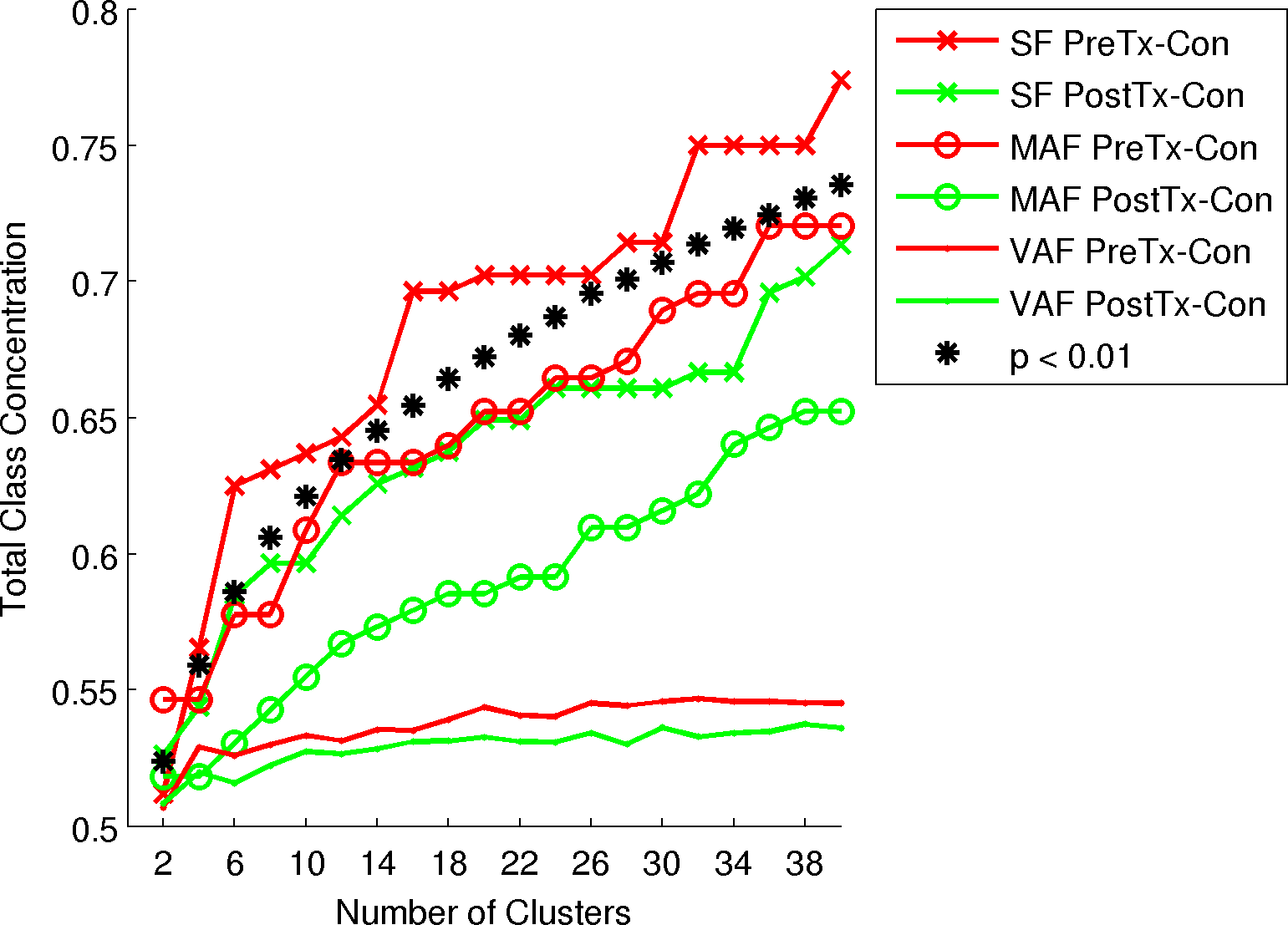}
\caption{The sensitivity of inter-subject clustering results for VAF, MAF and SF methods. The PreTx group is consistently more concentrated than the PostTx group for all methods, but only our method demonstrates the SF PreTx/Control clustering passing statistical significance.}
\label{fig:clustSensitivity}
\end{figure}

We first computed the concentration values for which RRDM passes the p $<$ 0.01 significance level; our SF features should ideally keep the total concentration of the PreTx/Control clustering over p $<$ 0.01, and the PostTx/Control clustering under p $<$ 0.01 to demonstrate that there are consistent differences from the Control subject-days in the PreTx group that are not present in the PostTx group after therapy. As shown in Figure \ref{fig:clustSensitivity}, the inter-subject class concentration increases as the number of clusters grows. The Vector Acoustic Features (VAF) perform worst, followed by the Mean Acoustic Features (MAF). The MAF PreTx-Con nears statistical significance. With our method (SF) PreTx-Con clusterings are significant at the 0.05 level on all but the very first clusters. We also have the PreTx and PostTx group separate when more than 5 clusters are used, and the separation passes the RRDM p $<$ 0.01 significance level. Specific clustering results for $n$ = 18 ($d$ = 0.116) are presented in Sections \ref{sec:precon} and \ref{sec:postcon}.\footnote{The distance between the the PreTx and PostTx concentrations was maximized in our method when 24 clusters were used (total class concentration difference = 0.124). However, $n=18$ minimized the number of clusters over the maximum concentration difference $d = conc_{class}(PreTx/Con) - conc_{class}(PostTx/Con)$, such that $d$ was not significantly lower than the absolute $\max\limits_{n}(d)$.} 
\vspace{-1.0em}

\section{Discussion and Related Work}
In this work we used unsupervised machine learning to analyze a novel clinical data set containing long-term time-series data. Prior work with ACC data has focused on targeted feature extraction for  supervised classification of subjects~\citep{ghassemi2014learning}. However, supervised learning is a poor method for detecting differences in the vocal behavior MTD patients, because people with MTD do not always speak in a disordered way, and there is no standard for labeling individual glottal pulses as disordered.

Our method differs from other recent work in three key ways: 1) We segment individual glottal pulses from the ACC signal rather than taking the traditional fixed-width frames; 2) We directly judge the relevance of a particular segmented item in our set by its morphology rather than using transforms derived from expert knowledge; and 3) We summarize a subject-day using a weighted sum over paired sets of morphological symbols and frequencies rather than a large set of features, or simple aggregates. From a clinical perspective, our results demonstrate that an ACC signal can be used to detect a difference in the vocal behavior of patients and controls. We also showed that vocal therapy has a measurable impact on patient behaviors.

Time-series symbolization~\citep{lin2003symbolic} and symbolic representation for time series based on sequence shape~\citep{patel2002mining} have previously been used to find time series motifs. Symbolization of segmented ECG data was used for supervised risk stratification~\citep{syed2007clustering} and assessing the clinical utility of expert-annotated heartbeats~\citep{li2012dynamic}. Unlike this prior work, we do not use symbolized distances as part of a supervised learning regime. Instead, we use these distances to represent using a set of prototypes based on density~\citep{lughofer2008extensions}.

More complex generative models have recently been developed for physiological problems, e.g., a multi-level latent model to learning individual and population level traits from clinical temporal data~\citep{saria2010learning} and incorporating multiple time series~\citep{saria2010discovering}. Symbolization is particularly attractive for developing clinical markers, since symbols are fast to extract and compare~\citep{syed2011unsupervised}, and variations in glottal pulse shape based on voice quality may be detectable with symbolization~\citep{laver1980phonetic}.



Our work is the first large scale study of vocal misuse based on long-term ambulatory data with over 100 million segments corresponding to glottal pulses from 253 subject-days of data.
The long-term goal of this multi-disciplinary project is to build a non-invasive ambulatory system that could be used to 1) diagnose voice disorders, 2) assess the impact of voice therapy, and 3) help facilitate the adoption of more normal vocal behaviors by providing biofeedback. 

\section{Acknowledgements}
We would like to thank Tristan Josef Naumann and David Kale for many helpful conversations, edits, and feedback. This research was funded in part by the Intel Science and Technology Center for Big Data, the National Library of Medicine Biomedical Informatics Research Training grant (NIH/NLM 2T15 LM007092-22), and the Voice Health Institute and the National Institutes of Health (NIH) National Institute on Deafness and Other Communication Disorders under Grants R33 DC011588 and F31 DC014412.

\bibliographystyle{abbrv}
\bibliography{Uncovering_Voice_Misuse_Symbolic_Mismatch_Marzyeh_Ghassemi_MLHC2016}

\begin{thebibliography}{23}
\providecommand{\natexlab}[1]{#1}
\providecommand{\url}[1]{\texttt{#1}}
\expandafter\ifx\csname urlstyle\endcsname\relax
  \providecommand{\doi}[1]{doi: #1}\else
  \providecommand{\doi}{doi: \begingroup \urlstyle{rm}\Url}\fi

\bibitem[Behrman et~al.(2003)Behrman, Dahl, Abramson, and
  Schutte]{behrman2003anterior}
Alison Behrman, Linda~D Dahl, Allan~L Abramson, and Harm~K Schutte.
\newblock Anterior-posterior and medial compression of the supraglottis: signs
  of nonorganic dysphoniaor normal postures?
\newblock \emph{Journal of Voice}, 17\penalty0 (3):\penalty0 403--410, 2003.

\bibitem[Buekers et~al.(1995)Buekers, Bierens, Kingma, and
  Marres]{buekers1995vocal}
R~Buekers, E~Bierens, H~Kingma, and EHMA Marres.
\newblock Vocal load as measured by the voice accumulator.
\newblock \emph{Folia phoniatrica et logopaedica}, 47\penalty0 (5):\penalty0
  252--261, 1995.

\bibitem[Demmink-Geertman and Dejonckere(2008)]{demmink2008neurovegetative}
L~Demmink-Geertman and PH~Dejonckere.
\newblock Neurovegetative symptoms and complaints before and after voice
  therapy for nonorganic habitual dysphonia.
\newblock \emph{Journal of Voice}, 22\penalty0 (3):\penalty0 315--325, 2008.

\bibitem[Ghassemi et~al.(2014)Ghassemi, Van~Stan, Mehta, Za{\~n}artu, Cheyne,
  Hillman, and Guttag]{ghassemi2014learning}
Marzyeh Ghassemi, J~Van~Stan, D~Mehta, Mat{\'\i}as Za{\~n}artu, H~Cheyne,
  R~Hillman, and J~Guttag.
\newblock Learning to detect vocal hyperfunction from ambulatory neck-skin
  acceleration features: Initial results for vocal fold nodules.
\newblock 61\penalty0 (6):\penalty0 1668–--1675, 2014.

\bibitem[Hsiung and Hsiao(2003)]{hsiung2003characteristic}
Ming-Wang Hsiung and Yu-Che Hsiao.
\newblock The characteristic features of muscle tension dysphonia before and
  after surgery in benign lesions of the vocal fold.
\newblock \emph{ORL; journal for oto-rhino-laryngology and its related
  specialties}, 66\penalty0 (5):\penalty0 246--254, 2003.

\bibitem[Laver(1980)]{laver1980phonetic}
John Laver.
\newblock The phonetic description of voice quality.
\newblock \emph{Cambridge Studies in Linguistics London}, 31:\penalty0 1--186,
  1980.

\bibitem[Li and Clifford(2012)]{li2012dynamic}
Q~Li and GD~Clifford.
\newblock Dynamic time warping and machine learning for signal quality
  assessment of pulsatile signals.
\newblock \emph{Physiological Measurement}, 33\penalty0 (9):\penalty0 1491,
  2012.

\bibitem[Lin et~al.(2003)Lin, Keogh, Lonardi, and Chiu]{lin2003symbolic}
Jessica Lin, Eamonn Keogh, Stefano Lonardi, and Bill Chiu.
\newblock A symbolic representation of time series, with implications for
  streaming algorithms.
\newblock In \emph{Proceedings of the 8th ACM SIGMOD workshop on Research
  issues in data mining and knowledge discovery}, pages 2--11. ACM, 2003.

\bibitem[Lughofer(2008)]{lughofer2008extensions}
Edwin Lughofer.
\newblock Extensions of vector quantization for incremental clustering.
\newblock \emph{Pattern Recognition}, 41\penalty0 (3):\penalty0 995--1011,
  2008.

\bibitem[MATLAB()]{matlabSigProcToolbox}
MATLAB.
\newblock \emph{Signal Processing Toolbox Release 2013b}.
\newblock The MathWorks, Inc.

\bibitem[Mehta et~al.(2012)Mehta, Zanartu, Feng, Cheyne, and
  Hillman]{mehta2012mobile}
Daryush~D Mehta, Matias Zanartu, Shengran~W Feng, Harold~A Cheyne, and Robert~E
  Hillman.
\newblock Mobile voice health monitoring using a wearable accelerometer sensor
  and a smartphone platform.
\newblock \emph{Biomedical Engineering, IEEE Transactions on}, 59\penalty0
  (11):\penalty0 3090--3096, 2012.

\bibitem[OECD()]{oecd}
OECD.
\newblock Oecd labour force statistics 2014.
\newblock \doi{http://dx.doi.org/10.1787/oecd_lfs-2014-en}.
\newblock URL \url{/content/book/oecd_lfs-2014-en}.

\bibitem[Ohlsson et~al.(1989)Ohlsson, Brink, and Lofqvist]{ohlsson1989voice}
Ann-Christine Ohlsson, Olle Brink, and Anders Lofqvist.
\newblock A voice accumulator—validation and application.
\newblock \emph{Journal of Speech, Language, and Hearing Research}, 32\penalty0
  (2):\penalty0 451--457, 1989.

\bibitem[Patel et~al.(2002)Patel, Keogh, Lin, and Lonardi]{patel2002mining}
Pranav Patel, Eamonn Keogh, Jessica Lin, and Stefano Lonardi.
\newblock Mining motifs in massive time series databases.
\newblock In \emph{Data Mining, 2002. ICDM 2003. Proceedings. 2002 IEEE
  International Conference on}, pages 370--377. IEEE, 2002.

\bibitem[Rantala and Vilkman(1999)]{rantala1999relationship}
Leena Rantala and Erkki Vilkman.
\newblock Relationship between subjective voice complaints and acoustic
  parameters in female teachers' voices.
\newblock \emph{Journal of Voice}, 13\penalty0 (4):\penalty0 484--495, 1999.

\bibitem[Roy et~al.(2005)Roy, Merrill, Gray, and Smith]{roy2005voice}
Nelson Roy, Ray~M Merrill, Steven~D Gray, and Elaine~M Smith.
\newblock Voice disorders in the general population: prevalence, risk factors,
  and occupational impact.
\newblock \emph{The Laryngoscope}, 115\penalty0 (11):\penalty0 1988--1995,
  2005.

\bibitem[Roy et~al.(2013)Roy, Barkmeier-Kraemer, Eadie, Sivasankar, Mehta,
  Paul, and Hillman]{roy2013evidence}
Nelson Roy, Julie Barkmeier-Kraemer, Tanya Eadie, M~Preeti Sivasankar, Daryush
  Mehta, Diane Paul, and Robert Hillman.
\newblock Evidence-based clinical voice assessment: A systematic review.
\newblock \emph{American Journal of Speech-Language Pathology}, 22\penalty0
  (2):\penalty0 212--226, 2013.

\bibitem[Saria et~al.(2010)Saria, Koller, and Penn]{saria2010discovering}
Suchi Saria, Daphne Koller, and Anna Penn.
\newblock Discovering shared and individual latent structure in multiple time
  series.
\newblock \emph{arXiv preprint arXiv:1008.2028}, 2010.

\bibitem[Saria et~al.(2012)Saria, Koller, and Penn]{saria2010learning}
Suchi Saria, Daphne~L Koller, and Anna~A Penn.
\newblock Learning individual and population level traits from clinical
  temporal data.
\newblock In \emph{In the Predictive Models in Personalized Medicine Workshop}.
  Twenty-Fourth Annual Conference on Neural Information Processing Systems,
  2012.

\bibitem[Stager et~al.(2003)Stager, Neubert, Miller, Regnell, and
  Bielamowicz]{stager2003incidence}
Sheila~V Stager, Rebecca Neubert, Susan Miller, Joan~Roddy Regnell, and
  Steven~A Bielamowicz.
\newblock Incidence of supraglottic activity in males and females: a
  preliminary report.
\newblock \emph{Journal of Voice}, 17\penalty0 (3):\penalty0 395--402, 2003.

\bibitem[Syed and Guttag(2011)]{syed2011unsupervised}
Zeeshan Syed and John~V Guttag.
\newblock Unsupervised similarity-based risk stratification for cardiovascular
  events using long-term time-series data.
\newblock \emph{Journal of Machine Learning Research}, 12:\penalty0 999--1024,
  2011.

\bibitem[Syed et~al.(2007)Syed, Guttag, and Stultz]{syed2007clustering}
Zeeshan Syed, John Guttag, and Collin Stultz.
\newblock Clustering and symbolic analysis of cardiovascular signals: discovery
  and visualization of medically relevant patterns in long-term data using
  limited prior knowledge.
\newblock \emph{EURASIP Journal on Applied Signal Processing}, 2007\penalty0
  (1):\penalty0 97--97, 2007.

\bibitem[Wang et~al.(2013)Wang, Mueen, Ding, Trajcevski, Scheuermann, and
  Keogh]{wang2013experimental}
Xiaoyue Wang, Abdullah Mueen, Hui Ding, Goce Trajcevski, Peter Scheuermann, and
  Eamonn Keogh.
\newblock Experimental comparison of representation methods and distance
  measures for time series data.
\newblock \emph{Data Mining and Knowledge Discovery}, 26\penalty0 (2):\penalty0
  275--309, 2013.

\end{thebibliography}

\appendix
\section*{Appendix A. Data and Pre-processing Details}
\label{sec:tablepop}
When we performed inter-subject comparisons, we used zero-mean, unit-variance amplitude normalization of ACC segments during the pre-processing phase. We did this because the habitual non-hyperfunctional volume level varies for individuals, especially in inter-gender comparisons. In intra-patient comparisons, however, we left segments scaled to daily dbSPL estimates, since we hypothesized that intra-patient comparisons would benefit from using the volume information to determine if an individual's own habitual loudness was affected by treatment. We provide a listing of data obtained for the cohort in Table~\ref{tab:recruitment}.

\begin{table}[h]
\caption{Subjects IDs are identified by their \textbf{N}ormal or \textbf{P}atient status (both \textbf{P}reTx and \textbf{P}ostTx), then by their gender (`M'ale or `F'emale), and finally by a unique three digit record number.}
\label{tab:recruitment}
\centering                                                                   
\begin{tabular}  
{p{1.5cm}|p{0.5cm}|p{0.6cm}|p{0.6cm}|p{0.5cm}|p{0.6cm}|p{0.6cm}}
& \multicolumn{3}{|c|}{Days} & \multicolumn{3}{|c}{Segments (Millions)}\\
\cline{2-7}
ID & \textbf{N} & \textbf{P}re & \textbf{P}ost & \textbf{N} & \textbf{P}re & \textbf{P}ost \\
\hline
A F023& 8 & 8 & 7 & 4.9 & 7.4 & 5.8\\
\cline{2-7}
B F027& 8 & 8 & 9 & 5.8 & 3.2 & 4.6\\
\cline{2-7}
C F040& 8 & 7 & 9 & 4.6 & 2.6 & 2.3\\
\cline{2-7}
D F048& 8 & 6 & 9 & 5.5 & 2.3 & 3.6\\
\cline{2-7}
E F052& 9 & 10 & 9 & 2.6 & 4.3 & 2.2\\
\cline{2-7}
F F064& 7 & 7 & 6 & 1.9 & 2.5 & 2.5\\
\cline{2-7}
G F069& 7 & 8 & 7 & 3.2 & 3.0 & 3.2\\
\cline{2-7}
H F071& 8 & 6 & 8 & 3.4 & 2.0 & 2.0\\
\cline{2-7}
I F100& 7 & 7 & 7 & 4.0 & 4.1 & 3.2\\
\cline{2-7}
J M035& 8 & 8 & 7 & 3.6 & 3.0 & 4.0\\
\cline{2-7}
K M074& 8 & 7 & 7 & 2.5 & 0.9 & 1.8\\
\hline
\end{tabular}
\end{table}

\end{document}